# Practical Uses of Belief Functions.

Philippe Smets
IRIDIA
Université Libre de Bruxelles
Brussels-Belgium

*The proof of the pudding is in the eating.*

### Abstract:
We present examples where the use of belief functions provided sound and elegant solutions to real life problems. These are essentially characterized by 'missing' information. The examples deal with 1) discriminant analysis using a learning set where classes are only partially known; 2) an information retrieval systems handling inter-documents relationships; 3) the combination of data from sensors competent on partially overlapping frames; 4) the determination of the number of sources in a multi-sensor environment by studying the inter-sensors contradiction. The purpose of the paper is to report on such applications where the use of belief functions provides a convenient tool to handle 'messy' data problems.

## 1. INTRODUCTION.

The models proposed to represent quantified uncertainty are based on probability, possibility or belief functions. They are complementary, as they cover different forms of uncertainty (Smets, 1998a). We consider here only those based on belief functions, in particular the hint model and the transferable belief model (TBM).

Showing one theory is better than another is often just impossible, as it requires a clear definition of 'better'. Better in what sense? Usually there are no definitive and absolute quality criteria, and most used criteria are either ad hoc or biased toward one theory.

What can then be done in order to compare models?
1. One can compare the underlying axioms and evaluate their respective adequacy and naturalness. Such axioms exist for each model, but there is no criterion that tells which one is really adequate.
2. One can compare the consequences of the various models and discard those leading to inadequate conclusions. But the conclusions to which they lead, are usually defendable, even when they don't agree, and there is no golden standard to select the 'winner'.
3. One can compare their abilities to solve small artificial delicate problems. But toy examples like the 3 prisoners, the 3 doors, the 'Peter, Paul and Mary Saga' don't lead to clear conclusions, as the merits of the solutions cannot be assessed definitively.
4. One can compare their usefulness in solving 'real' problems. This is what we try to do here by presenting some problems where the belief function approach was quite convenient.

A very important point when modeling uncertainty is to be clear about what is modeled, an obvious preliminary step that some skip too easily. Both probability and belief functions based models represented the weighted opinion of an agent that the actual world belongs to a given set of possible worlds, or equivalently that a given proposition is true in the actual world. Thus there is something called the 'actual world' and it has to be made clear what is really meant before even applying any model. In the actual world, as considered here, every proposition is either true or false: there is no fuzziness (belief on fuzzy events has been defined, but is not considered here).

It would be nice to compare the results obtained with belief functions with those one could obtain with a probabilistic approach. Some comparisons are presented here. But we realize the difficulty encountered when trying such a comparison. In fact belief functions are used when some of the data needed for a probability analysis are missing. If all such data were available, they should be fed into the belief function analysis... in which case the model reduces itself into the probability model. The whole argument about using belief functions centers on how the missing information is handled. Probabilists usually solve the problem by introducing some 'natural' assumptions like equi-probabilities, independence, or a modelization of the missing-ness. If these assumptions are 'close' to reality, the probability solution is often optimal, in which case using a belief functions approach is useless (we just hope that belief functions produce results almost as good as those obtained with probability functions). The real interest of the belief functions approach is to be found in its robustness to discrepancies between the assumptions and the reality. E.g., Appriou (1997) shows an analysis dealing with missile recognition where the belief functions approach was much more robust to these discrepancies than the probabilistic approach.

Besides, comparing the two approaches is difficult, as there are many methods to handle the missing information in probability theory. Bad results observed with probability functions could then be explained either by a weakness of the probability approach or by an inadequate choice of the method. Deciding which one applies is delicate.

We describe four examples where belief functions models produce nice and efficient solutions. These examples are more or less 'real life' examples. We nevertheless limit our presentation to simplified illustrations; their generalization to large-scale applications is immediate. These examples were essentially developed by Thierry Denoeux, University of Compiègne, France (Denoeux, 1995), by Johan Schubert, Royal Institute of Technology, Stockholm, Sweden (Schubert, 1995), by Justin Picard, Université de Neuchâtel, Switzerland (Picard, 1998) and by Janez, ONERA, Paris, France, (Janez, 1996). Another application of the TBM is presented in the paper 'Assessing the value of a candidate' by Dubois et al. (1999, see in this proceeding). The first application is presented in some detail and includes a comparison with probabilistic approach. For lack of space, the other three examples are just shortly described. Their authors describe in full detail the methods used and their interest.

This paper only reports on the use of belief functions in a few real applications recently developed. In depth comparisons with other methods are still missing. Benchmark exercises should be organized.

## 2. UNCERTAINTY AND BELIEF FUNCTIONS.

Shafer (1976) introduces a model to represent quantified beliefs based on so-called belief functions. Since, many new results have been obtained that we survey here. We neglect the computational issue: valuation based system, fast Möbius transform and approximation methods are detailed in (Gabbay and Smets, 1998-99, vol. 5).

In AI, Shafer's model was called 'Dempster-Shafer theory' (Gordon and Shortliffe, 1984). Unfortunately what this name covers varies widely from authors to authors (Smets, 1994). It can correspond to:
1. a lower probability model,
2. Dempster's model derived from probability theory (Dempster, 1967) and represented by the hints theory of Kohlas (Kohlas and Monney, 1995),
3. Shafer's model unrelated to probability theory (Shafer, 1977, 1992) and represented by the transferable belief model (Smets and Kennes, 1994, Smets 1997a, 1998).

The confusion between these interpretations explains most errors encountered in the literature where authors analyze Shafer's ideas.

### 2.1. The lower probability model.

Let a set $\Omega$, and let $\Pi$ be a set of probability functions defined on $\Omega$. The lower probability of a subsets A of $\Omega$ is defined as:
$P_*(A) = \min_{P \in \Pi} P(A)$ for every $A \subseteq \Omega$.
The $P_*$ function is also called the lower envelope of $\Pi$. Under certain weak constraints $P_*$ is a Choquet capacity monotone of order 2, and it might even be monotone of order infinite, in which case $P_*$ is a belief function. In all these cases, $P_*$ and $\Pi$ are in one-to-one correspondence.

There are at least two ways to get $\Pi$.
1. There exist a P function on $\Omega$ band the agent knows only that P belongs to $\Pi$. It can also be obtained by studying betting behaviors and calling $P_*(A)$ the maximal price the agent, called the player, would pay to a banker to enter a game where the player gets from the banker $1 if A occurs, and nothing otherwise (Walley, 1991). (The difference with the Bayesian definition is that the agent cannot be forced to be the banker).
2. Beliefs are represented by families of probability functions that can be defined through their lower envelop $P_*$ (Kyburg, 1987, Voorbraak, 1993).

An important issue when developing a model to represent beliefs is to explain its behavior when new pieces of evidence are introduced like in the conditioning process. In the first interpretation, the solution is obvious: every probability function in $\Pi$ is conditioned by Bayes rule on $A \subseteq \Omega$, and $P_{*A}$ is the lower envelop of this new set of probability functions (Jaffray, 1992). These results were often used to criticize Dempster's rule of conditioning, and indirectly Shafer's work. This comparison is inappropriate, as Dempster's rule of conditioning is *not* justified in this context. This approach is not further considered here after as belief functions are only marginally concerned.

### 2.2. The theory of hints.

Historically, the use of belief functions was initiated by Dempster while justifying fiducial inference (Dempster, 1967, 1968). Today its most developed model is the hint theory of Kohlas and Monney (1995). They assume Dempster's original structure $(\Omega, P, \Gamma, \Theta)$ where $\Omega$ and $\Theta$ are two sets, $P^\Omega$ is a probability measure on $\Omega$ and $\Gamma$ is a one-to-many mapping from $\Omega$ to $\Theta$. The set $\Theta$ is the set of possible answers to a question whose answer is unknown. One and only one element of $\Theta$ is the correct answer to the question. 'The goal is to make assertions about the answer in the light of the available information. We assume that this information allows for several different interpretations, depending on some unknown circumstances. These interpretations are regrouped into the set $\Omega$ and there is exactly one correct interpretation. Not all interpretations are equally likely and the known probability measure $P^\Omega$ reflects our information in that respect. Furthermore, if the interpretation $\omega \in \Omega$ is the correct one, then the answer is known to be in the subset $\Gamma(\omega) \subseteq \Theta$. Such a structure $H = (\Omega, P, \Gamma, \Theta)$ is called a hint... An interpretation $\omega \in \Omega$ supports the hypothesis H if $\Gamma(\omega) \subseteq H$ because in that case the answer is necessarily in H. The degree of support of H is defined as the probability of all supporting interpretation of H' (Kohlas and Monney, 1995, page vi).

The hints theory is similar to the probability of provability theory (Ruspini, 1986, Pearl, 1988, Smets, 1991a, 1993b), a theory that extends the domain of the probability functions from propositional logic to modal logic.

### 2.3. The transferable belief model.

Shafer (1976) proposes to quantify beliefs with belief functions, instead of probability functions as classically used for that purpose. He introduces the model and both Dempster's rule of conditioning that correspond to the Bayes conditioning rule, and Dempster's rule of combination that correspond to the probability aggregation rule.

From these ideas, we develop the TBM, a non-probabilistic model for representing the quantified beliefs held by an agent.

Let You denote the agent whose beliefs are considered, but it should be realized that an agent can also be a piece of equipment, a computer program, a sensor, etc... Subjectivity is not essential. Your beliefs manifest themselves at two mental levels: the credal level where beliefs are entertained and represented by belief functions, and the pignistic level where beliefs are used to act and represented by probability functions.

Suppose a frame of discernment $\Omega$ on which Your beliefs are considered. One world in $\Omega$ is the actual world, denoted $\omega_0$. You can only express the strength of Your opinion, Your belief, that $\omega_0$ belongs to this or that subset of $\Omega$. You allocate parts of Your belief to the fact that $\omega_0 \in A$ for every $A \subseteq \Omega$. The part of belief, denoted $m(A)$, given to $A \subseteq \Omega$ represents the part of Your belief that specifically supports the fact that $\omega_0 \in A$ and no set more specific than A. The total amount of belief, denoted $bel(B)$, that supports $\omega_0 \in B$ is obtained by adding the parts of belief $m(A)$ given to the sets A, $A \neq \emptyset$, $A \subseteq B$.

When You must take decisions, the belief held at the credal level, and represented by the basic belief assignment m defined on $\Omega$, induces a probability function at the 'pignistic' level, denoted BetP and also defined on $\Omega$. The transformation is called the pignistic transformation (Smets, 1989):

$$BetP(x) = \sum_{A \subseteq \Omega, x \in A} \frac{m(A)}{1-m(\emptyset)} \frac{1}{|A|} \text{ for every } x \in \Omega.$$

This probability function can be used in order to make decisions using expected utilities theory. Its justification is based on rationality requirements detailed in (Smets and Kennes, 1994)

The operational definition of a degree of belief is based on the agent's betting behaviors and its assessment is based on exchangeable bets just as it is done for subjective probabilities (Smets and Kennes, 1994)

The axiomatic of the TBM as a model to represent quantified beliefs is detailed in Smets (1993A, 1997a), (see also Wong et al., 1990). Axiomatic justification for combination rules is given in (Smets, 1990, Dubois and Prade, 1986, Hajek, 1992, Klawonn and Schwecke 1992). The generalization of the Bayesian Theorem to the TBM is presented in Smets 1978, 1993b). Decision process based on lower expectations are explained in Strat, (1990) Jaffray (1989), whereas Wilson (1993) studies the properties of the pignistic probabilities and Smets (1993c) examines what become the pignistic transformations in a dynamic decision making context.

Revision of beliefs by specialization are described in (Kruse and Schwecke, 1990, Klawonn and Smets, 1992) and several general combination rules have been developed (Dempster's rule of combination is hardly the only rule for combining two belief functions. There are many other rules, like the disjunctive rule of combination (Smets, 1993d), the $\alpha$-combination rules (Smets, 1997b, the cautious combination rules, etc...).

Principles of information content have been developed. Measures extending entropy measure are detailed in (Klir and Wierman, 1998), whereas we develop a measure adapted to the TBM (Smets, 1983). Principle of minimal commitment that states 'never give more support to a set that necessary' replaces the maximum entropy principle used in probability theory (Hsia, 1991)

Classically bel(A) quantifies 'I have good reason to believe A' and bel(A) is the strength of these 'good reasons'. In (Smets, 1995), we show how to represents concepts like 'good reasons *not* to believe', a concept similar in logic to the retraction à la Gärdenfors. Similarly, we can also express concepts like 'I still have some reasons to believe', and 'I still have some reasons not to believe'.

### 2.4. Future developments.

Many new developments have been achieved since Shafer's seminal work. Limiting oneself to the theory as presented in Shafer's book is no more acceptable. The distinction between the three interpretations for belief functions seems essential but it deserves further work to validate or invalidate it. Works on belief revisions - finding their nature and the adequate rules for representing their effect - are necessary. Dempster's rule of conditioning fits just one kind of revision. There are still open theoretical issues but it is obvious that real life applications are needed before the interest of the model can be assessed. In Europe there are already quite a few applications under development. They usually concern problems where some information essential for a probability approach is missing and cannot be obtained. The way belief functions can adequately represent partial or total ignorance is usually acknowledge. Belief functions are used for pattern recognition, multi-sensor data fusion, diagnosis... A nice property of belief functions is that only what is known is used.

## 3. THE TBM CLASSIFIER.

Discriminant analysis is probably the most classical tool used for classifying cases into one of several categories given the values of some measurement variables. Normally, we use a set of data, called the learning set (LS). For each case in LS, we know the values taken for each measurement variable and the classification variable that tells the class to which the case belongs. The classes are finite and unordered. Let $\Omega$ denote the set of possible classes: $\Omega = \{c_1, c_2,..., c_n\}$.

A learning set with N cases and p measurement variables is the set $\{(c_i, x_{1i}, x_{2i}, ... x_{pi}): i = 1, 2...N\}$ where $X_i$ is the 'name' of the i'th case, $c_i$ is the class to which $X_i$ belongs, and $x_{ji}$ is the value of the measurement variable j for $X_i$. The data of a new case, denoted $X_?$, is collected, but the class to which $X_?$ belongs, denoted $c_?$, is unknown. We want to predict the value of $c_?$ given the observed values of the measurement variables of $X_?$. Solutions to this problem are well established. One of them, called discriminant analysis, is fully described in most textbooks of statistics.

Let us now suppose that instead of the ideal learning set LS as described here above, we have a learning set PKLS where the classes of the cases are only partially known. For instance suppose we only know that case $X_1$ belongs either to $c_1$ or $c_2$ class, that case $X_2$ does not belong to class $c_1$, case $X_3$ belongs either to $c_2$ or $c_5$ or $c_7$ class ... Can we adapt the discriminant analysis method to such 'messy' data case? In fact we face a problem of 'partially supervised learning'. For some cases, classes are known as in the supervised learning approach, for some cases, class in completely as in the unsupervised approach. But here we also have all the cases where we know partially their class. Probabilistic solutions could be based on:
1. a Bayesian approach where we assess for each case a probability function that describes the class to which it belongs. We then allocate every case to a class (and get the probability to get that learning set), compute the needed parameters as in a supervised learning approach and average the results weighted by the probability of the learning sets.
2. a maximum likelihood approach where we estimate the unknown parameters, including the probability with which the case belong to a given class.
3. an adaptation of cluster analysis where partial constraints are introduced that represent the knowledge about the class to which each case belong.

Whatever method is used, the computational complexity is a serious problem and an adequate tuning of some parameters is not a small matter. The transferable belief model provides another approach that can handle elegantly and efficiently such a messy case. The method was invented by Denoeux (1995). We present results of the method – called the TBM classifier - and compare them with those obtained by the classical discriminant analysis applied to the same data base but using the exact value for the classes, a method that is then optimal. Details about these results are given in Denoeux (1995), Zouhal (1997), De Smet (1998).

### 3.1. Discriminant Analysis with Partially Known Classes

Let $pkc_i$ denote the subset of $\Omega$ that represents what we know about the class to which case $X_i$ belongs. The learning set PKLS is now the set $\{(pkc_i, x_{1i}, x_{2i}, ... x_{pi}): i = 1, 2...N\}$

Intuitively the method can be described by an anthropomorphic model. Each case $X_i$ in PKLS is considered as an individual. Let $c_{i0}$ denoted the true class to which $X_i$ belongs. All $X_i$ knows about $c_{i0}$ is that $c_{i0} \in pkc_i$ (Denoeux and Zouhal (1999) generalizes to the case where this knowledge is represented by a belief function or possibility function on $\Omega$). Then $X_i$ looks at the unknown case and expresses 'his' belief $bel_i$ about $c_?$. If $X_?$ is 'close' to $X_i$, $X_i$ would defend that $c_? = c_{i0}$. As all what $X_i$ knows about $c_{i0}$ is that $c_{i0} \in pkc_i$, then all what $X_i$ can express about case $X_?$ is that $c_? \in pkc_i$. If $X_?$ is not 'close' to $X_i$, $X_i$ cannot say anything about $c_{i0}$.

This description is formalized as follows. $X_i$ can only states: case $X_?$ belongs to the same set of classes as myself, what is represented by a belief function with $m_{i0}(pkc_i) = 1$. Let $d(X_i, X_?)$ be the 'distance' between $X_i$ and $X_?$. If $d(X_i, X_?)$ is small, then what $X_i$ stated is reliable, if $d(X_i, X_?)$ is large, it is not reliable, the largest $d(X_i, X_?)$, the less reliable. The impact of this reliability is represented by a discounting on $m_{i0}$ into $m_i$. So $m_i(pkc_i) = f(d(X_i, X_?))$ and $m_i(\Omega) = 1-f(d(X_i, X_?))$ where $f(d) \in [0,1]$ and is decreasing with d. Thus every case $X_i$ generates such a simple support function $bel_i$ on $\Omega$ that concerns the value of $c_?$.

Consider now what information $X_?$ collects. Case $X_?$ receives all these simple support functions $bel_i$, and combines them by Dempster's rule of combination into a new belief function bel on $\Omega$ that represents the belief

616    Smets

held by case $X_?$ about $c_?$ and induced by the collected belief functions $bel_i$:

$$bel_? = \oplus_{i=1...N} bel_i.$$

If a decision must be made on the value of $c_?$, we build the pignistic probability $BetP_?$ on $\Omega$ from $bel_?$ by the application of the pignistic transformation (described and justified in Smets and Kennes, 1994) and use the classical expected utility theory in order to take the optimal decision.

In the comparison study presented here after (and done by Y. De Smet 1998), we use the next solutions. Each measurement variable in PKLS is linearly re-scaled so that their 5th percentile is 0 and their 95th percentile is 1. So measurement variables share similar scales, and the method is robust to outliers.

For f, he uses: $f(d) = \max(1 - ad, 0)$ with $a>0$. More elaborated formulas were useless. For d, he uses the $D^2$ of Mahalanobis using a covariance matrix $\Sigma_i$ that depends on $X_i$ and which parameters are based on the cases in the neighborhood of $X_i$.

De Smet applied this approach to many sets of data. We present only six case studies. The quality criterion used in all comparisons is the classical PCC (percent of correct classification). The predicted class is always the class with the highest pignistic probability (the most probable class). Furthermore in every artificial case study, the pkc is never erroneous, i.e. the true class of $X_i$ belongs always to $pkc_i$.

### Case Study 1. Isosceles triangle, AB/AC/BC.

Suppose a two dimensional (p=2) trigroup classification problem with group labels A, B, C. Data in the three groups are normally distributed, their means are at the corner of an isosceles triangle with coordinate (0,0), (4,0) and (2,2), and the covariance matrix is the unit matrix. 200 cases are randomly generated in the 3 groups. Each set of 200 cases is split in two subsets of 100 cases, those in the first subsets having their label transformed into {A,B}, the others into {A,C}. The same is done for the other two groups. The learning set is made out of 20 A, 20 B, 20 C cases (randomly selected), the other case making the testing set. Table 1 presents the PCC obtained for 5 unrelated sets of data by the TBM-classifier (denoted TBM with pkc). For comparison purpose, we also present the PCC obtained by linear discriminant analysis, denoted DA, applied to the same data sets but using the true classes for the data in the learning set. Both methods produce similar results, an excellent result for what concerns the TBM-classifier. Indeed it only uses the partially known classes whereas the DA uses precisely known classes, a much richer information, and furthermore DA is the optimal method for these data as they satisfy exactly the requirements underlying the use of DA.

|   | PCC | | | | |
|---|---|---|---|---|---|
| 1 TBM with pkc | 93 | 93 | 92 | 89 | 94 |
| DA true class | 94 | 94 | 93 | 93 | 92 |
| 2 TBM with pkc | 94 | 94 | 93 | 93 | 89 |
| DA true class | 96 | 95 | 96 | 95 | 95 |
| 3 TBM with pkc | 85 | 81 | 84 | 86 | 80 |
| DA true class | 88 | 86 | 89 | 85 | 86 |

**Table 1:** For each case studies 1 to 3, PCC obtained in five experiments with the TBM classifier using partially known classes and with discriminant analysis (DA) using the true classes.

### Case Study 2. Collinear, A/B/AC-BC.

Data are generated as in case study 1, except the means are collinear, located at (-3,0), (3,0) and (0,0). The pkc of the A cases is A, and the pkc of the B cases is B (there is no missing information for these two sets of data). The C cases where all classified are either {A,C} or {B,C}. The difficulty comes from the fact that the C cases are essentially located between the A and B cases. The results in table 2 support the conclusions of case study 1.

### Case Study 3. 5 classes, in $R^5$

We use p=5 and 5 classes, denoted A, B...E. The mean of group 1 is located on the first axis at $2\sqrt{2}$, for group 2, on the second axis at $2\sqrt{2}$, etc... The covariance matrix = I. The pkc are build as follows. Suppose a case in group 2 as indicated by the vector (0, 1, 0, 0, 0). Then for every 0 in the vector, we toss a fair coin: if heads we put a 1, if tails we leave the 0. Then the 1's in the resulting vector indicate those subsets included in the pkc of that case. So if the end vector is (1, 1, 0, 1, 0), the pkc is {A, B, D}. The learning set is made out of 30 cases from each class. Even though the knowledge about the class was quite poor, the TBM-classifier provided results (see table 3) almost as good as the discriminant analysis approach applied on perfectly known classes (again optimal here). That the PCC with the TBM-classifier are lower is normal, no miracle can be expected, the TBM-classifier used a very imprecise information, and a method using more information should provided better results.

| $\sigma^2$ | TBM Classifier | Linear Discrimination |
|---|---|---|
| 10 | 84 | 85 |
| 15 | 78 | 79 |
| 20 | 75 | 77 |
| 25 | 75 | 77 |
| 30 | 73 | 74 |
| 50 | 65 | 65 |

**Table 2:** Case study 4: impact of large variances.

### Case Study 4. Triangle, No pkc.

In order to see if the TBM-classifier behaves well when the classes are precisely known, we use the isosceles triangle of case study 1, with side length = 10, and a covariance matrix $\sigma^2 I$ with $\sigma^2$ varying from 10 to 50. The linear discriminant is against the optimal method in such a case. Table 2 shows that the TBM-classifier performs as well as the linear discriminant method, whatever $\sigma^2$.

**Case Study 5. Collinear, No pkc.**
As in cased study 2, we use 3 groups with 100 cases per group and the means are collinear located at (10,0) (20,0) (30,0). With the covariance matrices $\Sigma_A = \Sigma_B = \Sigma_C = 50.I$, both the TBM and the linear discriminant classifiers produce PCC of 69%. When $\Sigma_A = \Sigma_C = 10.I$ and $\Sigma_B = \begin{pmatrix} 10 & 0 \\ 0 & 1 \end{pmatrix}$, the PCC are 73% for the TBM classifier and 51% for the linear discriminant classifier (which condition for optimality are unsatisfied here, but in real life it is not obvious to realize it and linear discriminant classifiers are often applied in such cases). The nice conclusion is that the TBM classifier is robust against such bad data.

**Case Study 6: Real Data, no pkc.**
We move now to real data where the classes are precisely known, just to show that the TBM-classifier behaves similarly to some of its competitors (De Caestekere, 1997). She uses a 1-Nearest Neighbor classifier, a multi-layers perceptron method, a prototype Nearest Neighbor method and we apply to the same data sets the TBM-classifier. these classifiers are applied to the too famous Iris data set, the diabetes data set and the wave data sets presented in De Caestekere. Their major characteristics are presented in table 3. Data where used as given, or with added white noises. Results of the TBM-classifier (table 4) are as good as those obtained with the other three approaches that are usually acknowledged as being among the best.

| Data Set | Dimension | Classes | Training | Test |
|---|---|---|---|---|
| Iris 1 | 4 | 3 | 75 | 75 |
| Iris 2 | 4 | 3 | 30 | 120 |
| Wave | 21 | 3 | 300 | 5000 |
| Diabetes | 5 | 3 | 71 | 74 |

**Table 3:** Major characteristics of the data sets used in table 4.

| Data | 1-NN | MLP | PNN | TBM |
|---|---|---|---|---|
| Iris 1 | 94.2 | 96.8 | 93.8 | 90.8 |
| Iris 2 | 94,7 | 96,0 | 95,6 | 95,7 |
| Iris 1B | 79,3 | 81,5 | 82,0 | 78,0 |
| Iris 2B | 80.7 | 83.0 | 83.0 | 82.8 |
| Diabetes | 97.3 | 98.0 | 99.1 | 93.5 |
| Wave | 76.4 | 84.9 | 83.8 | 82.5 |

**Table 4:** PCC observed on data of table 3 with four classification methods.

The conclusions of the comparisons based on the six case studies illustrated here (and many others done buy De Smet) are:
4. when the classes are precisely known, the TBM-classifier performs almost as well as the classical classifiers.
5. the TBM-classifier can be used in cases where classes are partially known, in which cases performance is still very good.

That the TBM-classifier can be applied in the case of partially known classes provides its real interest, as such messy data situations can hardly be handled with the classical tools as available today.

**3.2. Partial knowledge is real life.**

It seems the TBM-classifier provides a nice tool, but does it fill a real need. The answer is affirmative. Real life hardly complies with the perfect knowledge usually required by classical statistical tools. Real life is messy data, not idealized data as one hopes for. As an example, consider the clinician who collect during the 1980's the data from 300 hundred patients suffering from a given disease Dx. In the 80's such patients were classified as A or B? Then as science advances, a new category C is described for patients with disease Dx. So during the 90's, our clinician collect 200 data classified as A, B or C. The clinician comes to you and asks for a computerized classifier. How to handle the first 300 cases, the A cases were in fact A or C, and the B cases were B or C, and their exact classes cannot be re-assessed. Are you going to throw away the 300 cases as useless... With the TBM-classifier, you can proceed with all the 500 cases, whereas a plain statistical analysis would have serious troubles with the learning set.

Suppose another clinician who collects data in 4 classes denoted A, B, C and D.
Then regulation or knowledge changes and these cases are supposed to be classified into three classes, denoted X, Y and Z, where all A cases are now X cases, all D cases are now Z cases, and the B cases turn out to be either X or Y cases, and the C cases turn out to be either Y or Z cases. How to handle the old database? This is exactly what is illustrated by case study 2.

Suppose a disease with 3 forms denoted A, B and C and three clinicians, denoted a, b, and c. Dr. a can only differentiate between A and not-A cases. The A cases are treated by Dr. a, the not-A cases are send to the hospital H. The same scenario holds for Dr. b where A is replaced by B, and for Dr. c where A is replaced by C. So at hospital H, the only cases they treat where classified as not-A, not-B or not-C, and there is no way to find out what was the exact class of these patients (as if the only available information is the name of the sending doctor). This is exactly what is illustrated by case study 1.

This shows that 'artificial' examples are not that artificial.

It would be interesting to compare these results when classes are only partially known with other techniques. The major difficulty is of course in the construction of alternate methods based on probability theory as extra assumptions will have to be fed into the models, and the quality of the results will strongly depends on the adequacy of these assumptions. In practice the user is not aware of this adequacy before using the classifier.

## 4. PAS FOR INFORMATION RETRIEVAL.

Justin Picard (1998) applies the Probabilistic Argumentation Systems, denoted PAS, (Kohlas and Haenni, 1999), an adaptation of the hint model, and a generalization of the ATMS of Laskey and Lehner (1989) to a problem of information retrieval, using in particular the CACM collection (3204 documents, 50 queries). Let a query Q, a set of documents $D_i$ and the citation links between them, denoted Citing($D_i$, $D_j$). The citation link reflects the idea that if a document $D_i$ is relevant to Q and cites $D_j$, then $D_j$ is probably also relevant to Q. For each document $D_i$, he introduces an assumption $a_i$. When assumption $a_i$ holds then $D_i$ is relevant to Q. When assumption $a_i$ does not hold, then nothing can be concluded about the relevance of $D_i$ to Q. To assess the probability $\alpha_i$ that assumption $a_i$ holds, he uses the rank of $D_i$ as provided by the search engines present on the Web. He fits $\alpha_i$ by a logistic regression.

$$\alpha_i = \frac{\exp^{-2.42\ln(\text{rank}_i)+1.11}}{1+\exp^{-2.42\ln(\text{rank}_i)+1.11}}$$

Picard then builds the PAS for modeling document relationship like the one in figure 5. The numerous cycles in the graph should be enhanced; they do not create any problem when using the PAS methodology. Another assumption $I_{ij}$ is introduced. If there is a citation link between $D_i$ and $D_j$, if $D_i$ is relevant to Q and if $I_{ij}$ holds then document $D_j$ is also relevant to Q. If $I_{ij}$ does not hold, noting can be concluded about the relevance of $D_j$ to Q that would result from the citation link between them. He accepts that the probability $\lambda$ that $I_{ij}$ holds does not depend on the documents involved. The fitted value for $\lambda$ is .2644.

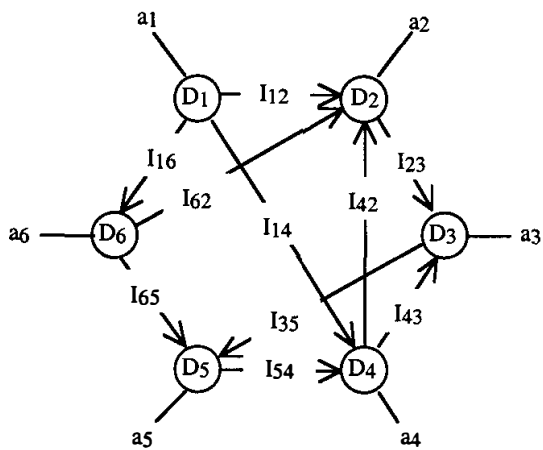

The figure represents 'graphically a PAS for a collection of six documents having some document relationships.

Rules are represented by arrows and assumptions by white dotted circles (e.g., $a_1 \rightarrow D_1$, $a_2 \rightarrow D_2$,...$D_1 \wedge I_{12} \rightarrow D_2$...) One can see that the support of $D_6$ (or any document) corresponds to all existing path from any a priori assumption to $D_6$: the support of $D_6$ is $a_6 \vee (a_1 \wedge I_{16})$. $D_6$ can thus be "proven" either by the retrieval system (argument $a_6$) or by document $D_1$ through the link from $D_1$ to $D_6$ (argument $a_1 \wedge I_{16}$).

Document $D_4$ illustrates how PAS deal with cycles. There are links from $D_4$ to $D_3$, $D_3$ to $D_5$ and $D_5$ to $D_4$. Even if there is a cycle here, evidence is counted only once: recall that the support of $D_4$ is the disjunction of all arguments for which $D_4$ becomes true. $a_5 \wedge I_{54}$ is one such argument. Since it implicitly contains $(a_5 \wedge I_{54} \wedge I_{43} \wedge I_{35} \wedge I_{54})$ ( = $a_5 \wedge I_{54} \wedge I_{43} \wedge I_{35}$) which would correspond to the cycle $D_4$-$D_3$-$D_5$-$D_4$, this last argument is not considered. Anyway, the algorithm for determining the symbolic support eliminates cycles.' (Quotations are from Picard, 1998).

## 5. SENSORS ON PARTIALLY OVERLAPPING FRAMES.

Suppose a sensor $S_1$ that has been trained to recognize A objects and B objects and another sensor $S_2$ that has been trained to recognize B objects and C objects (like A = airplanes, B = helicopters and C = rockets). Sensor $S_1$ never saw a C object, and we know nothing on how $S_1$ would react if to a C object. Beliefs provides by $S_1$ are always on the frame of discernment {A, B}. The same holds for $S_2$ with A and C permuted. A new object X is presented to the two sensors. Both sensor $S_1$ and $S_2$ express their belief $m_1$ and $m_2$, the first on the frame {A, B}, the second on the frame{B, C}. How to combine these two beliefs on a common frame $\Omega$ = {A, B, C}? Solutions have been proposed in Janez (1996).

Solutions are based on the next constraint. If both $m_1$ and $m_2$ are conditioned on {B}, and combined by Dempster's rule of combination (unnormalized), the resulting belief function should be the same as the one obtained after 'combining' the original $m_1$ and $m_2$ on [A, B, C}, and conditioning the result on {B}. The problem is of course how to 'combine' m1 and m2. The original Dempster's rule of combination is inadequate as it requires that both belief functions are defined on the same frame of discernment, what is not the case here.

A general solution is as follows. Let $\Omega_1$ and $\Omega_2$ be the frame of discernment of $m_1$ and $m_2$, respectively. Let $\Omega = \Omega_1 \cap \Omega_2$. For all $A \subseteq \Omega_1 \cup \Omega_2$, let $A_1 = A \cap \Omega_1$, $A_2 = A \cap \Omega_2$, $A_0 = A \cap \Omega$, and

$$m(A) = \frac{m_1(A_1)}{m_1[\Omega](A_0)} \frac{m_2(A_2)}{m_2[\Omega](A_0)} (m_1[\Omega] \oplus m_2[\Omega])(A_{12})$$

where $m_1[\Omega]$ and $m_2[\Omega]$ are the basic belief assignments obtained by conditioning $m_1$ and $m_2$ on $\Omega$. In table 5, we illustrate the computation. We have $m_1[B]\oplus m_2[B](B) = (.1+.3)*(.7+.1) = .32$. This mass is distributed on $\{B\}$, $\{A, B\}$, $\{A, C\}$ and $\{A, B, C\}$ according to the next ratios: $(.1/.4).(.7/.8)$, $(.3/.4).(.7/.8)$, $(.1/.4).(.1/.8)$, and $(.3/.4).(.1/.8)$. In this example the first sensor supports that X is an A, whereas the second claims that X is a B. If X had been a B, how comes the first did not say so? So the second sensor is probably facing an A and just states B because it does not know what an A is. So we feel that the most plausible solution is X = A, what is confirmed by $BetP_{12}$ being the largest for A: $BetP_{12}(A) = .455$.

Just to enhance the simplicity of the belief function solution, we examine what this problem would be when expressed within probability theory. Suppose two sensors $S_1$ and $S_2$. Sensor $S_1$ generates a probability function on $\{A, B, C\}$, denoted $P*_1$, but we only know $P_1 = P*_1(.|\{A,B\})$, the value of $P*_1$ after conditioning it on $\{A, B\}$. The same holds for sensor $S_2$ with $P_2 = P*_2(.|\{B,C\})$. Aggregate $P_1$ and $P_2$ in order to derive a probability function on $\{A, B, C\}$. The major issue is on how to reconstruct $P*_1$ and $P*_2$ from $P_1$ and $P_2$. It means how to 'de-condition' a probability function on $\Omega$ when all you know is the result of its conditioning on some strict subset of $\Omega$. Suppose $\Omega = \{a, b, c, d\}$ and you know $P(\{a\}|\{a,b\})$. What would be $P(\{c\})$ and $P(\{c,d\})$? There are infinitely many solutions. Introducing the maximum entropy principle leads to $P(\{c\}) = P(\{d\}) = .25$ and $P(\{a\}) = .5 * P(a|\{a,b\})$. Such a solution is strongly linked to the insufficient reason principle and suffers of all its weaknesses.

| $\Omega$ | $m_1$ | $m_2$ | m | pl | BetP |
|---|---|---|---|---|---|
| {A} | .6 | | .00 | .92 | 455 |
| {B} | .1 | .7 | .07 | .32 | 190 |
| {C} | | .2 | .00 | .72 | 355 |
| {A, B} | .3 | | .21 | 1 | |
| {A, C} | | | .68 | .93 | |
| {B, C} | | .1 | .01 | 1 | |
| {A, B, C} | | | .03 | 1 | |

**Table 5.** Basic belief assignment $m_1$ and $m_2$ on the two partially overlapping frames, with their combination m and its related plausibility and pignistic probability functions.

## 6. ANALYZING CONTRADICTION AND THE NUMBER OF SOURCES.

Suppose a piece of equipment has failed. We collect data from four sensors $S_1$, $S_2$, $S_3$ and $S_4$. Each sensor produces a belief function on the set of possible component that might have failed. Table 6 presents a highly simplified example where each sensor produces a simple support function pointing toward one component. $S_1$ and $S_2$ both point toward component $C_1$, whereas $S_3$ and $S_4$ point toward component $C_2$. If the four sources $S_1$ to $S_4$ were highly reliable, you would conclude that both $C_1$ and $C_2$ are broken. Indeed if only one has failed, the source are contradictory, whereas if two components have failed, results are coherent if $S_1$ and $S_2$ report on one broken component and $S_3$ and $S_4$ report on a second broken component.

| $\Omega$ | $m_1$ | $m_2$ | $m_3$ | $m_4$ |
|---|---|---|---|---|
| $C_1$ | .7 | .8 | | |
| $C_2$ | | | .6 | .9 |
| $\Omega$ | .3 | .2 | .4 | .1 |

**Table 6.** The simple support functions generates by the four sensors on the frame of discernment $\Omega = \{C_1, C_2, \ldots C_n\}$.

How do we translate this problem into belief functions language? The solution is obtained by considering the mass $m(\emptyset)$ given to $\emptyset$ that may be positive in the transferable belief model. When applying Dempster's rule of combination to two basic belief assignments $m_1$ and $m_2$, the result is given by:

$$m_{12}(A) = \sum_{X,Y\subseteq\Omega, X\cap Y=A} m_1(X)m_2(Y) \text{ for all } A \subseteq \Omega.$$

We do not normalize the resulting basic belief assignment $m_{12}$, $m(\emptyset)$ is among the computed masses and it does not have to be 0 like in Shafer's original presentation. The mass $m(\emptyset)$ quantifies the amount of contradiction between the various sources of belief functions.

Schubert (1995) has proposed a strategy to decide the number of events under consideration by the various sensors producing the several collected belief functions. He analyses $m(\emptyset)$ and finds the association between sensors and events that somehow brings the total conflict to an acceptable level.

Suppose the data of table 6. If there is only one broken component the four sensors are speaking about the same event. The contradiction computed after combining the four basic belief assignments is 90, what reflect an enormous conflict between the four sources. If there is two broken components, then some sources might speak about one, the other about the second. So we split the four sensors into two groups, compute what is the contradiction within each group, and sum these contradictions. For instance, suppose sensors $S_1$, $S_2$ and $S_3$ speak about one component, then the contradiction is 0.56, whereas there is no contradiction for sensor 4. Total contradiction is thus 0.56. Now if we consider that sensor $S_1$ and $S_3$ speak about one component, whereas $S_2$ and $S_4$ speak about the other, the total contradiction is 1.14. Contradiction completely disappears when $S_1$ and $S_2$ are grouped as reporting on one component, and $S_3$ and $S_4$ on the second. This result fits with common sense analysis of the data. In real life applications, the basic belief assignments are usually quite elaborated, and finding an adequate grouping is not obvious. The technique of 'peeling' the mass given to the empty set (to the contradiction) is nevertheless still applicable. The level of 'tolerable contradiction' is itself determined by the analysis of the conflict present in the given belief functions (and obtained by the use of the canonical decomposition of the belief functions (Smets, 1995)).

The mass m(Ø) acts in fact as a measure of discrepancy between several belief functions. The proposed algorithm leads to grouping sources which belief functions are 'close' to each other. In probability theory using cross-entropy or chi-square coefficients can achieve this purpose. Comparisons between these approaches are not available (as far as we know). The advantage of the belief function approach resides in the well-founded nature of the approach. The mass m(Ø) is part of the transferable belief model, whereas the cross-entropy, the chi-square and the likes need always extra assumptions in order to justify their use.

| Groups | | Conflict | | |
|---|---|---|---|---|
| $G_1$ | $G_2$ | $G_1$ | $G_2$ | total |
| 1234 | - | .90 | - | .90 |
| 123 | 4 | .56 | .00 | .56 |
| 124 | 3 | .85 | .00 | .85 |
| 134 | 2 | .67 | .00 | .67 |
| 234 | 1 | .77 | .00 | .77 |
| 12 | 34 | .00 | .00 | .00 |
| 13 | 24 | .42 | .72 | 1.14 |
| 14 | 23 | .63 | .48 | 1.11 |

**Table 7.** Masses m(Ø) computed from the belief functions included in each group when considering two objects.

## 7. FINAL REMARKS.

We have presented a few 'real' life problems where the use of belief functions is interesting. These problems are characterized by the presence of some missing information that are needed to apply the probability approach. Probabilists normally try to obtain some 'estimation' of the values for these missing data and apply their model with these data (using sensitivity analysis in order to check the robustness of their conclusions to reasonable variations around the guessed 'estimation'). Usually the belief function models do not need such assumptions and is well adapted to work with the information as really available. This power comes from the ability of belief functions to represent any form of uncertainty: full knowledge, partial ignorance, total ignorance (and even probability knowledge). Probability functions do not have such expressiveness power. Equi-probability is not full ignorance, it is already a quite precise form of knowledge.

In practice, the major interest of the belief function approach as presented here comes from its robustness (Appriou, 1997, Picard, 1998). When the 'estimations' of the missing data are close to what they are in reality, the probability model is normally perfect. But once differences increase between the 'estimations' and reality, the probability models deteriorate much faster then the belief function models. It is amazing to note that long before belief functions had been introduced, Hüber (1973) had developed robust methods in statistics and his results have some similarity with those of the belief functions approach.

The computational issue is real but as shown in these examples, it seems manageable. No serious studies are available on that issue. We feel that the computational complexity will be similar to those encountered in probability theory, but of course brute force approaches must be avoided. E.g., in the Schubert's example, it is obvious that the sensor clustering will not be achieved by testing every partition, and that some stepwise approaches have to be used. The fact that belief functions are defined on the power set, whereas probability functions are defined on the set has often been used as an argument against the use of belief functions. Theoretically the argument is correct, but in practice situations will hardly be so complex and there are even cases where the complexity is smaller with belief functions. In any case approximations will be used.

**Acknowledgement.** Research work has been partly supported by the ESPRIT IV, Working Group FUSION funded by a grant from the European Union.